\documentclass{spie}

\usepackage{cite}

\usepackage{times} 
\usepackage{indentfirst} 

\usepackage{amsmath}
\usepackage{amsfonts}
\usepackage{mathtools}
\usepackage{multirow}
\usepackage{xcolor}
\usepackage{doi}
\usepackage{url}
\usepackage{xkeyval}

\title{MIDV-2019: Challenges of the modern mobile-based document OCR}

\author{Konstantin Bulatov\supit{1,2}, Daniil Matalov\supit{1, 2}, Vladimir V. Arlazarov\supit{1, 2}
  \skiplinehalf
  \normalsize 
  \supit{1} Federal Research Center ``Computer Science and Control'' of Russian Academy of Sciences, Moscow, Russia; \\
  \supit{2} Smart Engines Service LLC, Moscow, Russia
}

\begin{document}

\maketitle

\begin{abstract}
    Recognition of identity documents using mobile devices has become a topic of a wide range of computer vision research. The portfolio of methods and algorithms for solving such tasks as face detection, document detection and rectification, text field recognition, and other, is growing, and the scarcity of datasets has become an important issue. One of the openly accessible datasets for evaluating such methods is MIDV-500, containing video clips of 50 identity document types in various conditions. However, the variability of capturing conditions in MIDV-500 did not address some of the key issues, mainly significant projective distortions and different lighting conditions. In this paper we present a MIDV-2019 dataset, containing video clips shot with modern high-resolution mobile cameras, with strong projective distortions and with low lighting conditions. The description of the added data is presented, and experimental baselines for text field recognition in different conditions.

    The dataset is available for download at \url{ftp://smartengines.com/midv-500/extra/midv-2019/}.
  
\end{abstract}

\section{Introduction}

Usage of smartphones and tablet computers for solving business process optimization problems in enterprise systems, as well as processes in government systems, lead to a new development turn for computer vision systems operating on mobile devices. The increased interest in implementing corporate workflow management using mobile documents processing, and the necessity of entering document data in uncontrolled conditions elevate the requirements for document recognition, entry, and analysis systems which use mobile devices \cite{mollah_design, google-doc-10}.

The images obtained using mobile cameras have a range of specific properties and distortions, such as low resolution (especially for low-end smartphones and tablet computers), insufficient or inconsistent lighting, blur, defocus, highlights on reflective surfaces of the objects of interest, and others \cite{arl-small-scale-cameras}. Such properties increase the requirements for mobile optical recognition systems and brings necessity to the development of new methods and algorithms which would be more robust against such distortions. This particularly concerns the models and methods of optical recognition of objects in camera-based environments, autonomous methods which can work in isolated mobile computational systems (and thus dealing with constrained computational power) \cite{7994570, Takhirov:2016:EAC:2934583.2934615, Yanai:2016:EMI:2964284.2967243}, and methods for analyzing video stream input in real time \cite{8270252, vestnik_integration}. To facilitate the research on these topics adequate open datasets should be created and maintained.

A particular interest in the field of mobile computer vision systems is given to the task of identity document recognition~\cite{8283074, SMART_IDREADER_ICDAR}. Automatic entry of data from identity documents is used in such industries as fintech, banking, insurance, travel, e-government, and in such processes as user identification and authentication, KYC/AML (Known Your Customer~/~Anti-Money Laundering) procedures and others. Computer vision problems which are associated with automatic identity document entry using mobile devices, include:
\begin{enumerate}
    \item Determining the document class, type, subtype, or country of issue;
    \item Document boundaries detection in an image, or document page segmentation from the background;
    \item Per-field document segmentation and layout analysis;
    \item Personal photo detection or facial features extraction;
    \item Optical character recognition, capturing and recognition of text fields and properties of the document;
    \item Video stream analysis in real time;
    \item Image quality estimation;
    \item Security features detection, optical variable devices analysis (holograms, dynamic color embossing, etc.);
    \item Other related tasks.
\end{enumerate}

An important issue which comes up in relation to research and scientific publications on the topic of identity document processing is the availability of datasets. Identity documents contain sensitive personal information, so storing, transmitting or otherwise make this data public is impossible. In order to facilitate research in some of the topics mentioned above, the MIDV-500 dataset was introduced \cite{midv500-arxiv}. The dataset contained video clips of 50 identity documents with different types. Since it's impossible to create a public dataset of valid and authentic identity documents, the dataset contained mostly ``sample'' or ``specimen'' documents which could be found in WikiMedia and which were distributed under public copyright licenses. And thus, although the variability of the documents used in the dataset is comparatively low, the target objects featured in this dataset shared the common identity document features.

\begin{figure}[ht!]
  \centering
  \includegraphics[width=\textwidth]{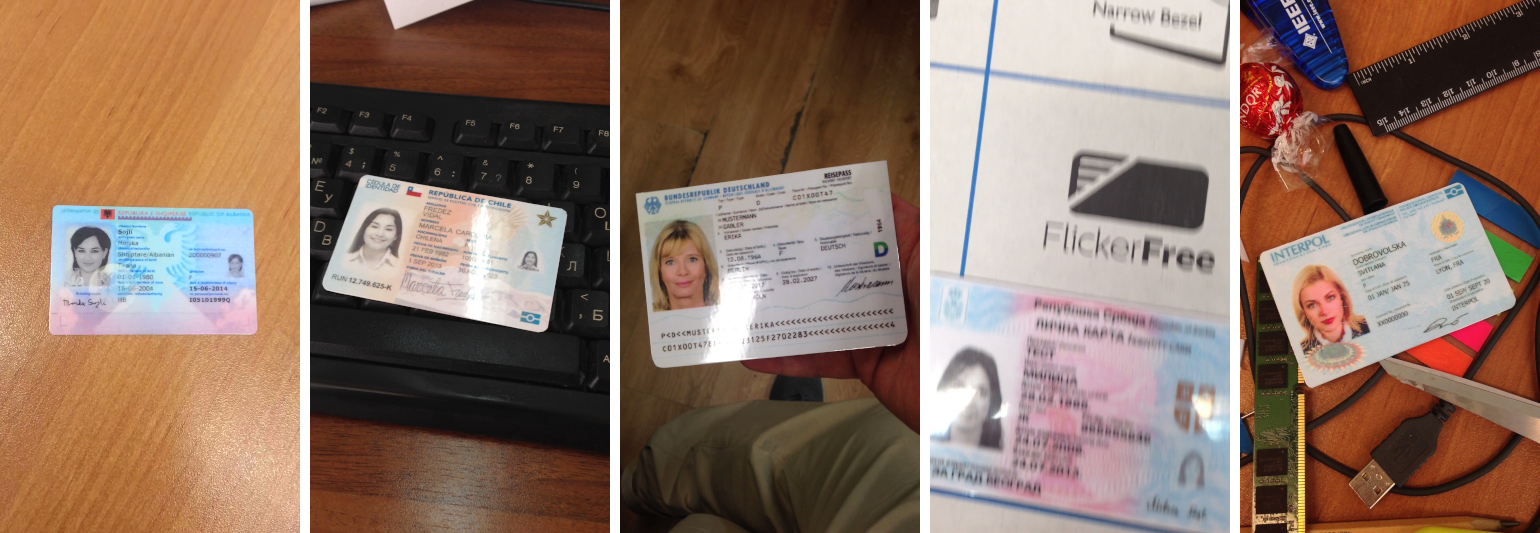}
  \caption{Conditions of the original MIDV-500 dataset \cite{midv500-arxiv}. From left to right: ``Table'', ``Keyboard'', ``Hand'', ``Partial'', and ``Clutter''}
  \label{fig:midv500_samples}
\end{figure}

The MIDV-500 dataset contained 500 video clips, 10 clips per document type. The clips were captured using Apple iPhone 5 and Samsung Galaxy S3 (GT-I9300), the smartphone models which could already be considered obsolete by the time the dataset was published. However, the increasingly common usage of identity document recognition in various business or government processes implies the need to support a wide range of devices, from cheap low-end devices to the ``flagship'' models. Using each of the two smartphone models, each document was shot in five distinct conditions: ``Table'', ``Keyboard'', ``Hand'', ``Partial'', and ``Clutter''. The ``Table'' condition represented the simplest case with the document laying on the table with homogeneous surface texture. The ``Keyboard'' represented the case when the document lays on various keyboards thus making it harder to utilize conventional edge detection methods, because of the background cluttered with straight edges and text. The ``Hand'' condition represented the case of a hand-held document. The ``Partial'' condition had some frames when the document is partially or completely hidden outside the camera frame. Finally, the ``Clutter'' condition had the background intentionally cluttered with random objects. The conditions represented in MIDV-500 dataset had some diversity regarding the background (``Table'', ``Keyboard'', ``Hand'', and ``Clutter'') and the positioning of the document in relation to the capturing process (``Partial''), however they did not include variation in lighting conditions, significant projective distortions, or a variation in the quality of the camera. Example images of every condition presented in MIDV-500 are illustrated in Figure \ref{fig:midv500_samples}.

The MIDV-500 dataset has been used to perform research of methods for document type recognition using similarity metric aimed at high classification precision and robustness against projective distortions \cite{lynchenko2019document}, for evaluation of image quality assessment methods and their impact on the document recognition system performance \cite{chernov2019application}, for analyzing the per-frame text field recognition results combination in a video stream \cite{vestnik_integration} and construction a stopping rule for text string recognition in a video stream \cite{Bulatov2019}.

Even though modern smartphones and tablet computers have camera modules with substantially higher quality, and their computational power have increased drastically, recognition of identity documents in images or a video stream in uncontrolled capturing conditions remains a scientific and technological challenge. In many use cases the documents are captured not by trained personnel, but remotely by document holders who are likely to be performing such capture very rarely and do not have any information about the processing algorithms involved. Thus, such conditions as low scene lighting, high projective distortions, and other complications, lead to a demand of sophisticated processing techniques and methods, even if input images are captured with high-end mobile devices. 

In this paper we present an extension to the MIDV-500 dataset, called MIDV-2019, which consists of video clips of the 50 original identity documents, but shot with more complex conditions and using high-end smartphone cameras. The two complex conditions targeted in this dataset extension are low lighting and strong projective distortions. The prepared extension is aimed to provide a platform for creating and evaluation of new methods and algorithms, designed to operate in challenging environments.

\section{Dataset description} \label{sec:dataset}

\begin{figure}[ht!]
  \centering
  \includegraphics[width=\textwidth]{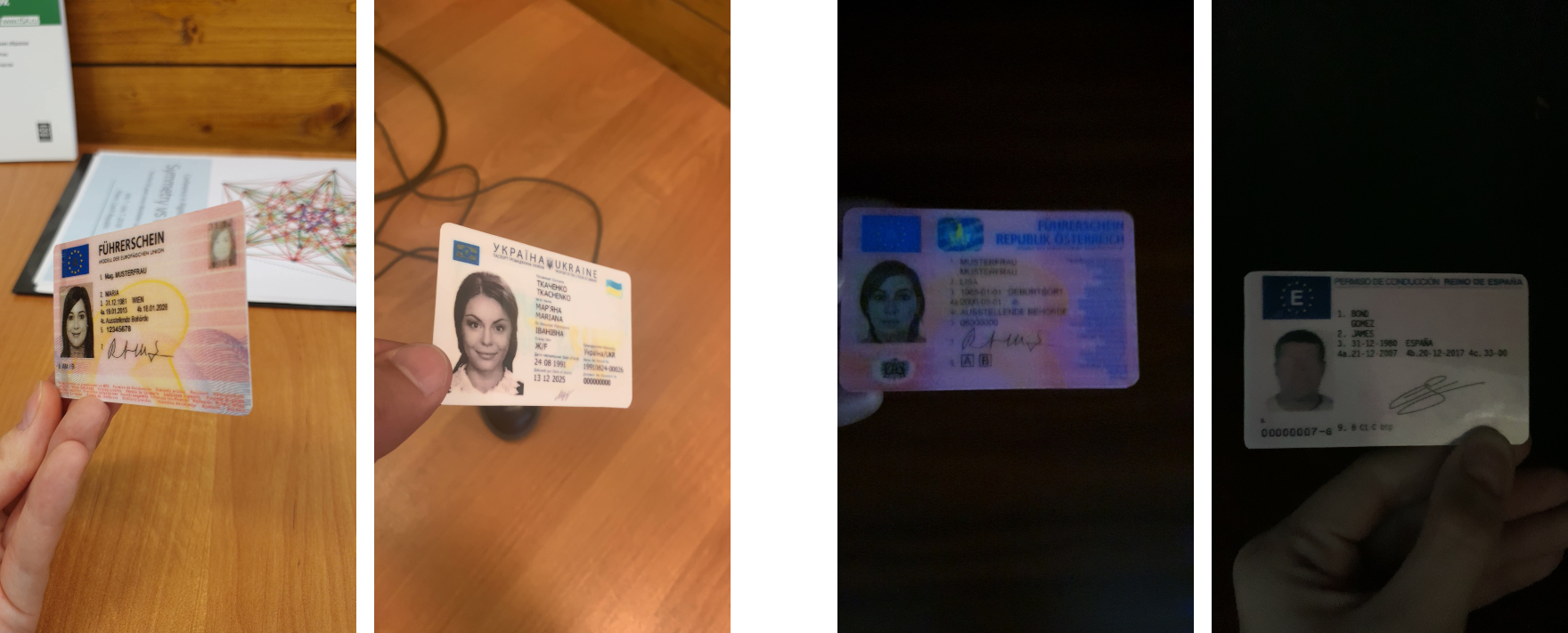}
  \caption{Conditions of the MIDV-2019 dataset. ``Distorted'' condition (left) and ``Low-lighting'' condition (right)}
  \label{fig:midv2019_samples}
\end{figure}

As with the MIDV-500 dataset, the new dataset MIDV-2019, presented in this paper, contains video clips of 50 different identity document types, which includes 17 ID cards, 14 passports, 13 driving licences and 6 other identity documents of different countries. The same printed samples which were used as a source of MIDV-500 dataset were also used to prepare MIDV-2019. For each printed document video clips were recorded under two different capturing conditions and using two mobile devices, thus obtaining 4 new video clips per document (200 new video clips in total). New clip identifiers are described in Table \ref{tbl:new_conditions}. Sample images of the added conditions are presented in Figure \ref{fig:midv2019_samples}.

\begin{table} \normalsize
\caption{Clip types added in MIDV-2019}
\label{tbl:new_conditions}
\begin{center}
\begin{tabular}{|p{0.08\textwidth}|p{0.6\textwidth}|}
    \hline
    Identifier & Description \\
    \hline
    DG & ``\underline{D}istorted'' -- documents were shot with higher projective distortions, videos were captured using Samsung \underline{G}alaxy S10 (SM-G973F/DS) \\ \hline
    
    DX & ``\underline{D}istorted'' -- documents were shot with higher projective distortions, videos were captured using Apple iPhone \underline{X}S Max \\ \hline

    LG & ``\underline{L}ow-lighting'' -- documents were shot in very low lighting conditions, video were capturing using Samsung \underline{G}alaxy S10 (SM-G973F/DS) \\ \hline
    
    LX & ``\underline{L}ow-lighting'' -- documents were shot in very low lighting conditions, video were capturing using Apple iPhone \underline{X}S Max \\ \hline
    
\end{tabular}
\end{center}
\end{table}

The first new capturing condition introduced in the MIDV-2019 dataset is the ``Distorted'' condition (clips ``DG'' and ``DX''), in which the documents were shot with strong projective distortions. The requirement for the document recognition systems to be able to operate in uncontrolled condition sometimes lead to the users capturing the documents with high projective distortions~--~for example, to avoid highlights on the reflective surfaces of the document. The methods which perform preliminary document detection and localization try to rectify the document image prior to processing, however it is important to have a highly distorted dataset of samples in order to assess the limits to the applicability of such methods. For methods which perform text segmentation and recognition without prior rectification \cite{HE2019107026, ElBahi2019}, and, specifically, the text components of identity documents, such as machine-readable zones \cite{SkoriukinaMRZ}, such capturing conditions may provide a valuable reference.

\begin{figure}[ht!]
  \centering
  \includegraphics[width=0.5\textwidth]{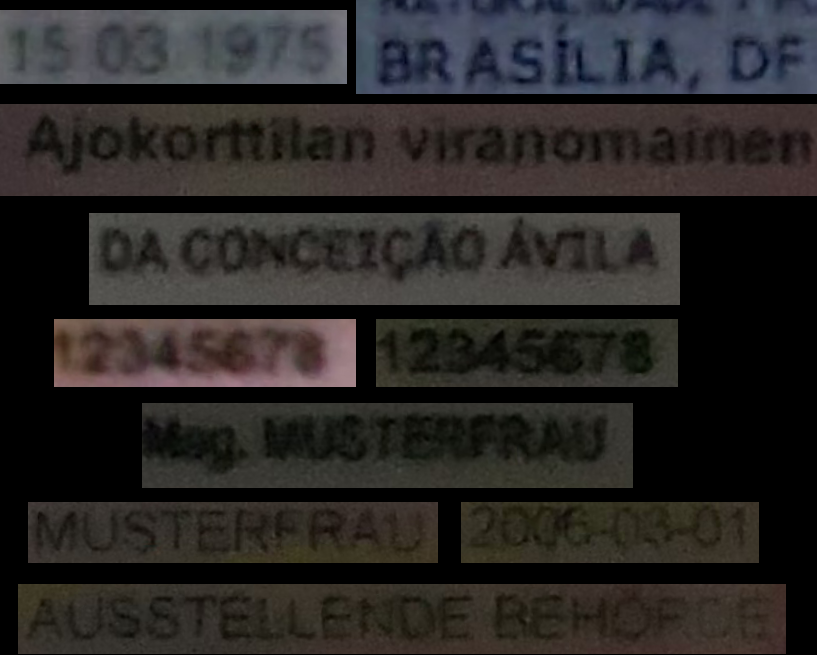}
  \caption{Examples of text fields cropped from ``LG'' and ``LX'' clips}
  \label{fig:low_lighting_fields}
\end{figure}

Perhaps the most significant challenge added in the MIDV-2019 dataset is the clips shot in a low-lighting conditions without flash. Such use cases as checking the identity documents in a long-distance travel, using mobile systems to enter identity document data by law enforcement officials, and others, sometimes require the ability to recognize documents with very low ambient light. In the images thus obtained the text is still visible and could be discerned by a human, however modern OCR systems struggle with this task. This is the primary reason for adding the clips ``LG'' and ``LX'', which represents the ``Low-lighting'' condition, to the MIDV-2019 dataset. The examples of text field images cropped from frames captured in low-lighting condition are presented in Figure \ref{fig:low_lighting_fields}.

All clips were shot in Ultra HD resolution (2160x3840). Each video were at least 3 seconds in duration and the first~3 seconds of each video were split with 10 frames per second. As in the MIDV-500 dataset, for each frame the ideal coordinates of the document's boundaries were annotated by hand, and if the corners of the document were not visible on the frame, the corresponding coordinate points were extrapolated outside the frame. The provided document coordinates combined with an ideal template segmentation ground truth, which is provided with the original MIDV-500, allow to crop document elements, such as text fields, from each frame, and perform evaluation of algorithms for face detection, text field recognition, as well as full document detection, classification, and segmentation.

\section{Evaluation baselines}

In order to provide basic baselines for future experiments based on the presented MIDV-2019 dataset we performed text field recognition evaluation using an open-source recognition system Tesseract v4.1.0 \cite{tesseract-paper}. As in the original paper presenting MIDV-500, four field groups were analyzed: numeric dates, document numbers, machine-readable zone lines, and Latin name components (which contain only Latin characters with no diacritical marks). Only the frames on which the document boundaries laid fully inside the frame were considered. To be consistent with MIDV-500, all fields were cropped with the resolution of 300 DPI (achieved using known physical dimensions of all document types present in the dataset), and for each field a margin was allowed with width equal to 10\% of the minimal dimension of the field's bounding box.

Table \ref{tbl:fields_total} lists the number of field images thus extracted from both MIDV-500 clips and from MIDV-2019 clips, grouped by the capturing condition. The first five rows of the table represent conditions in MIDV-500 dataset (``Table'', ``Keyboard'', ``Hand'', ``Partial'', and ``Clutter''). The last two rows represent conditions from the new MIDV-2019 dataset (``Distorted'' and ``Low-lighting'').

\begin{table}
  \caption{Number of analyzed text field images per field group and clip type}
  \label{tbl:fields_total}
  \resizebox{\textwidth}{!}{
    \begin{tabular}{|l||r|r|r|r|r||r|r|}
      \hline
      \multirow{2}{*}{\parbox{0.12\columnwidth}{Field group}} & \multicolumn{5}{c||}{MIDV-500} & \multicolumn{2}{c|}{MIDV-2019} \\
      
      \cline{2-8}
      &
      \parbox{0.11\columnwidth}{TS, TA} & 
      \parbox{0.11\columnwidth}{KS, KA} & 
      \parbox{0.11\columnwidth}{HS, HA} & 
      \parbox{0.11\columnwidth}{PS, PA}  &
      \parbox{0.11\columnwidth}{CS, CA} & 
      \parbox{0.11\columnwidth}{DG, DX} &
      \parbox{0.11\columnwidth}{LG, LX} 

      \\
      \hline
      
      Numeric dates & $4884$ & $4230$ & $3864$ & $796$ & $3961$ & $5122$ & $5390$ 
      
      \\
      \hline
      
      Document numbers & $2555$ & $2234$ & $2003$ & $435$ & $2102$ & $2648$ & $2841$

      \\
      \hline
      
      MRZ lines & $1504$ & $1232$ & $1072$ & $154$ & $1134$ & $1600$ & $1764$

      \\
      \hline
      
      Latin names & $4258$ & $3706$ & $3317$ & $740$ & $3566$ & $4350$ & $4676$
      
      \\
      \hline
    \end{tabular}
  }
\end{table}

In Table \ref{tbl:fields_accuracy} the text field recognition accuracy is presented for all aforementioned conditions and grouped by the type of the text field. The comparison of the recognized and correct values was case-insensitive and there were no distinction between the Latin letter ``O'' and the digit ``0''. While given the fixed recognition system the absolute values of the recognition accuracy is of a lesser interest, the main distinction which should be noted is between the capturing conditions. From the seven analyzed conditions the highest quality is achieved in the simplest case -- the one with document laying on the table (``TA'', ``TS'') with homogeneous background and with the smallest projective distortion. Even though the clips ``TA'' and ``TS'' were shot with older smartphone models and with Full HD resolution (1080x1920), the recognition precision for images taken in those conditions turned out to be higher than that of clips ``DG'' and ``DX'', which were shot with higher projective distortions but with Ultra HD resolution (2160x3840).

\begin{table}
  \caption{Text field recognition accuracy (percentage of correctly recognized fields) per field group and per clip type. Recognition performed using Tesseract v4.1.0, comparison was case-insensitive and the letter ``O'' and the digit ``0'' were treated as identical}
  \label{tbl:fields_accuracy}
  \resizebox{\textwidth}{!}{
    \begin{tabular}{|l||r|r|r|r|r||r|r|}
      \hline
      \multirow{2}{*}{\parbox{0.12\columnwidth}{Field group}} & \multicolumn{5}{c||}{MIDV-500} & \multicolumn{2}{c|}{MIDV-2019} \\
      
      \cline{2-8}
      &
      \parbox{0.11\columnwidth}{TS, TA} & 
      \parbox{0.11\columnwidth}{KS, KA} & 
      \parbox{0.11\columnwidth}{HS, HA} & 
      \parbox{0.11\columnwidth}{PS, PA}  &
      \parbox{0.11\columnwidth}{CS, CA} & 
      \parbox{0.11\columnwidth}{DG, DX} &
      \parbox{0.11\columnwidth}{LG, LX} 

      \\
      \hline
      
      Numeric dates & $47.420$ & $37.967$ & $42.107$ & $24.497$ & $32.517$ & $41.976$ & $6.106$ 
      
      \\
      \hline
      
      Document numbers & $46.458$ & $34.467$ & $38.842$ & $26.207$ & $36.108$ & $36.405$ & $6.864$

      \\
      \hline
      
      MRZ lines & $8.910$ & $5.844$ & $3.078$ & $1.948$ & $5.467$ & $6.625$ & $0.283$

      \\
      \hline
      
      Latin names & $55.636$ & $40.799$ & $54.658$ & $30.946$ & $41.615$ & $54.805$ & $13.045$
      
      \\
      \hline
    \end{tabular}
  }
\end{table}

By far the lowest text field recognition accuracy can be seen on the ``Low-lighting'' clips ``LG'' and ``LX''. Even if shot with high resolution and on modern smartphones, the task of document recognition in such conditions is still a clear challenge and should be addressed by the community. It should be noted that while the recognition accuracy on the ``Low-lighting'' clips is very low, the accuracy ordering by text field group is mostly the same as for the other conditions.

\section{Conclusion}

In this paper we presented the dataset MIDV-2019 containing video clips of identity documents captured using modern smartphones in low lighting conditions and with higher projective distortions. The paper presents experimental baselines for text field recognition for different capturing conditions and for different field groups presented in the dataset, and the reported result show that the text field recognition in low lighting is still a very challenging problem for modern mobile recognition systems. With the added data the MIDV-500 dataset is expanded by 40\%.

Authors believe that the provided dataset will serve as a valuable resource for document recognition research community and lead to more high-quality scientific publications in the field of identity documents analysis, as well as in the general field of computer vision.

\acknowledgements

This work is partially financially supported by Russian Foundation for Basic Research (projects 17-29-03170 and 17-29-03236). Source images for MIDV-2019 dataset are obtained from Wikimedia Commons. Author attributions for each source image are listed in the description table at \url{ftp://smartengines.com/midv-500/documents.pdf}.

\bibliographystyle{spiebib}
\bibliography{bibliography}


\clearpage



\end{document}